\documentclass{bioinfo}
\copyrightyear{2022} \pubyear{2022}

\access{Advance Access Publication Date: Day Month Year}
\appnotes{Manuscript Category}

\usepackage{enumerate}
\usepackage{paralist}
\usepackage{url}
\usepackage{xcolor}
\usepackage{natbib}
\usepackage{hyperref}
\usepackage{comment}
\usepackage{multirow}
\usepackage{booktabs}

\usepackage[russian,english]{babel}

\begin{document}
\selectlanguage{english}
\firstpage{1}

\subtitle{Data and text mining}

%\title[Nested Entities in Biomedical Abstracts]{A Unified   Scheme for Annotating Nested  Entities in Biomedical Abstracts} 
\title[Nested Entities in Biomedical Abstracts]{NEREL-BIO: A Dataset of  Biomedical Abstracts Annotated  with Nested Named Entities %\textcolor{red}{Suggest removing Russian since there is also English}
} 

\author[Loukachevitch et al.]{Natalia Loukachevitch\textsuperscript{1},Suresh Manandhar\textsuperscript{2}, Elina Baral\textsuperscript{2}, Igor Rozhkov\textsuperscript{1},   Pavel Braslavski\textsuperscript{3,4}, Vladimir Ivanov\textsuperscript{5}, Tatiana Batura\textsuperscript{6},  Elena Tutubalina\textsuperscript{7,8}}

\address{
\textsuperscript{1}Lomonosov Moscow State University, Moscow, Russia ~~\\
\textsuperscript{2}Madan Bhandari University of Science and Technology, Nepal~~\\
\textsuperscript{3}Ural Federal University, Yekaterinburg, Russia\\
\textsuperscript{4}HSE University, Moscow, Russia ~~ 
\textsuperscript{5}Innopolis University, Innopolis, Russia \\
\textsuperscript{6}A.P. Ershov Institute of Informatics Systems, Novosibirsk, Russia\\
\textsuperscript{7}Artificial Intelligence Research Institute, Moscow, Russia ~~
\textsuperscript{8}Sber AI, Moscow, Russia 
}

\corresp{$^\ast$To whom correspondence should be addressed.}

\history{Received on XXXXX; revised on XXXXX; accepted on XXXXX}

\editor{Associate Editor: XXXXXXX}

\abstract{
\textbf{Motivation:} 
This paper describes NEREL-BIO  -- an annotation scheme and corpus of PubMed abstracts in Russian and smaller number of abstracts in English. NEREL-BIO extends the general domain dataset NEREL~\citep{NEREL}   by introducing domain-specific entity types. 
NEREL-BIO annotation scheme covers both general and biomedical domains making it suitable for domain transfer experiments. %Like its predecessor, NEREL-BIO provides annotation for nested named entities and relations into nested named entities. 
NEREL-BIO provides annotation for nested named entities as an extension of the scheme employed for NEREL.
Nested named entities may cross entity boundaries to connect to shorter entities nested within longer entities, making them harder to detect. %NEREL-BIO is annotated with relations at three levels: within nested named entities, intra- and inter- sentences.  We provide benchmark evaluation of current state-of-the-art methods in all three tasks. 
%\keywords{Named entity recognition \and Nested entities \and Relation extraction \and Nested relations \and Entity linking}
%\subclass{ 68T35 \and 68T50} 
%.
\\
\textbf{Results:} NEREL-BIO contains annotations for 700+ Russian and 
%. It also contains annotations for 
100+ English abstracts. All English PubMed annotations have corresponding Russian counterparts. Thus, NEREL-BIO comprises the following specific features: annotation of nested named entities, it can be used as a benchmark for cross-domain (NEREL $\rightarrow$ NEREL-BIO) and cross-language (English $\rightarrow$ Russian) transfer. We experiment with both transformer-based sequence models and machine reading comprehension (MRC) models and report their results. 
\\
\textbf{Availability:} The dataset is freely available at \url{https://github.com/nerel-ds/NEREL-BIO}.
%We apply current state-of-the-art models for nested named entity recognition such as Machine Reading Comprehension (MRC). %transformer-based 
%\citep{devlin2018bert} 
%models \. %We provide experimental results using BioBERT \citep{lee2019biobert, bluebert}. %We extend both with entity linking.  
%We also experiment with domain transfer by initially training  models in the general domain using the NEREL corpus with subsequent fine-tuning on NEREL-BIO. `%\pb{don't think we need that many references in the abstract}
}

%We present a neural approach for medical concept normalization of diseases and drugs. Our two-stage approach is based on Bidirectional Encoder Representations from Transformers (BERT). 
%In the training stage, we optimize the relative similarity of mentions and concept names from a terminology via triplet loss. In the inference stage, we obtain the closest concept name representation in a common embedding space to a given mention representation. We performed a set of experiments on a dataset of abstracts and a real-world dataset of trial records with interventions and conditions mapped to drug and disease terminologies. The latter includes mentions associated with one or more concepts (in-KB) or zero (out-of-KB, \textit{nil} prediction). Experiments show that our approach significantly outperforms baseline and state-of-the-art architectures. Moreover, we demonstrate that our approach is effective in knowledge transfer from the scientific literature to clinical trial data. \\
%\textbf{Availability:} We make code and data freely available at \urlstyle{tt}\url{hidden\_during\_review\_process}\\
%\textbf{Contact:} \email{artur@insilico.com}, \email{kudrin@insilico.com}, \email{elena@insilico.com}\\
%\textbf{Supplementary information:} Supplementary data are available at \textit{Bioinformatics}
%online.
%}

\maketitle

\section{Introduction}

The lack of richly annotated training datasets is a well-known challenge in developing biomedical entity extraction systems. The majority of existing datasets and named entity recognition (NER) methods have been designed for capturing flat (non-nesting) mention structures over coarse entity type schemes. Moreover, the annotated entities in these corpora are limited to the most common entity types such as drugs/chemicals and diseases \citep{leaman2009enabling,gurulingappa2010empirical,van2012eu,wei2016assessing}.

GENIA~\citep{kim2003genia} is a widely studies corpus for biomedical named entity recognition in English consisting of 2{,}000 PubMed abstracts with more than 400{,}000 words and almost 100{,}000 annotations for biological terms. The annotated abstracts are devoted primarily to biological reactions concerning transcription factors in human blood cells. 47 entity types organized in taxonomy were annotated. The annotation includes nested and fragmented (non-continuous) entities. Yet, only 17\% of the entities in the GENIA corpus are nested within another entity \cite{katiyar2018nested}. \cite{mohan2018medmentions} describes MedMentions corpus of 4{,}392 PubMed abstracts, which is annotated with $21$ entity types, including disorders, anatomical structures, chemicals, and also some general concepts such as organizations, population groups, etc. However, we chose to annotate the most specific concept in texts without any overlaps in mentions.

Recent work has shown an increase in interest in nested entity structure on general-domain data on various languages, including English \citep{ringland2019nne}, Russian \citep{NEREL}, Thai \citep{buaphet2022thai}, and Danish \citep{plank2020dan+}. 
Most research so far, including for Russian, focused on newswire data. As for the biomedical and clinical domain in Russian, there are several datasets of clinical texts or drug-related user reviews with flat entities \cite{tutubalina2021russian,nesterov2022ruccon}. A recent work on a Russian medical language understanding benchmark \citep{blinov2022rumedbench} includes the RuDReC corpus \citep{tutubalina2021russian} for NER. However, these corpora ignore nested entities, like ``pain in the head'' being both a disease and an anatomy entity. To encourage the development of state-of-the-art information extraction systems aimed at providing more comprehensive coverage of biomedical concepts, we decided to construct a large nested named entity dataset \textbf{NEREL-BIO} over Russian PubMed abstracts. All entity mentions, including nested structures with up to six layers of depth, are manually annotated.

\begin{figure}[t!]
    \centering
    \includegraphics[width=0.4\textwidth, height=25mm] 
    {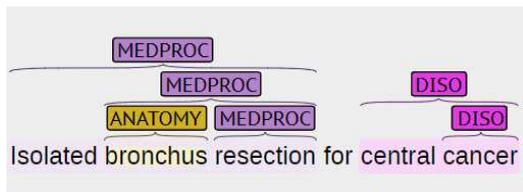}
    \caption{Example of nested named entities in NEREL-BIO. Internal entities ``bronchus'', ``bronchus resection'', ``resection'' are available in UMLS. Longer entities ``isolated bronchus resection'' and ``central cancer'' are important for establishing appropriate relations. %\pb{move Figure to page 2?}
    }
    \label{fig:example}
\end{figure}

Fig~\ref{fig:example} presents an example of nested named entities in NEREL-BIO. It discusses ``isolated bronchus resection for central cancer'' and provides the results of surgical treatment in these specific conditions. Entities ``bronchus'', bronchus resection'', ``resection'' are included in the Unified Medical Language System (UMLS) \citep{bodenreider2004unified}, while ``isolated bronchus resection'' and ``central cancer'' are not.  Nested entity annotations create a basis for establishing relations between correct (longer) entities, as well as linking internal entities to equivalent UMLS concepts.\\

The main contributions of our work are summarized as follows:
\begin{enumerate}
    \item We present NEREL-BIO, a new biomedical dataset for nested NER in Russian with a smaller corpus in English.
    \item We evaluate BERT-based Machine Reading Comprehension (MRC) and sequence models \citep{li2020unified,shibuya2020nested} for biomedical nested NER. 
    \item To promote further cross-lingual research, we annotate a subset of 100+ English abstracts in translation from Russian using the same annotation scheme.
\end{enumerate}

\section{Data collection and annotation}

\begin{table}[t!]
\centering
\processtable{Statistics of NEREL-BIO. \label{tab:generalstats}} 
{\begin{tabular}{@{}l|r|r|r|r@{}}
\hline
\textbf{Collection} & \textbf{\#Doc} & \textbf{\#Entities}& \textbf{\#Non-zero}&\textbf{\#Entities with}\\
&&& \textbf{entity types}&\textbf{freq > 50}\\
\hline
Abstracts in Russian & 766 & 66,888& 41&32\\
Abstracts in English & 105 & 10,651& 32&18\\

\hline
\end{tabular}}{}
\end{table}

\begin{table}[t!]
\centering
\processtable{Ten most frequent diseases in NEREL-BIO (translated from Russian) \label{tab:diseases}} 
{\begin{tabular}{@{}lr@{}}
\hline
\quad \textbf{Disease} & \textbf{\#Mentions}\\
\hline
\quad Tumor & 198\\
\quad Diabetes & 112\\
\quad Pain & 111\\
\quad Cancer & 101\\
\quad Tuberculosis & 100\\
 \quad Arterial hypertension & 88\\
\quad Infection & 86\\
\quad Stroke & 85\\
\quad Cardiac ischemia & 82\\
\quad Alzheimer's disease & 69\\

\hline
\end{tabular}}{}
\end{table}

\begin{table*}[t]
\centering
\processtable{UMLS semantic groups and NEREL-BIO entity types\label{tab:entity_types}}
%{\begin{tabular}{|l|l|l|l|c|}
{\begin{tabular}{@{}llllrr@{}}\toprule

%\hline
\textbf{UMLS semantic types} & \textbf{NEREL-BIO types} 
&\textbf{UMLS CUI}& \textbf{Initial Dataset}&\textbf{Freq (Rus)}&\textbf{Freq (Eng)}\\
\midrule
%\hline
A1 \textbf{Physical Object}&&&&\\
A1.1 Organism&LIVB&C0029235&NEREL-BIO&730&69\\
A1.1.3.1.1.4.1 Human&PERSON&C0086418&NEREL&7258&907\\
A1.2 Anatomical structure & ANATOMY&C1384516&NEREL-BIO&8992&1842\\
A1.2.3.5 Gene or Genome&GENE&C0010325&NEREL-BIO&376&2\\
A1.3 Manufactured Object&&&\\
&AWARD&C0004446&NEREL&2&0\\
&MONEY&C087090&NEREL&35&0\\
&WORK\_OF\_ART&NO&NEREL&0&0\\
A1.3.1 Medical device&DEVICE&C0699733&NEREL-BIO&661&40\\
&PRODUCT&C1514468& NEREL&85&4\\
A1.4.1 Chemical &CHEM&C0220806&NEREL-BIO&5049&1109\\
A1.4.3 Food&FOOD&C0016452&NEREL-BIO&135&10\\
\midrule
A2 \textbf{Conceptual entity}  &&&&\\
A2.1 Idea or Concept
&IDEOLOGY&C0815330&NEREL&0&0\\
&RELIGION&C0035039&NEREL&0&0\\
A2.5 Language&LANGUAGE&C0023008&NEREL&5&0\\
A2.1.1 Temporal&DATE&C0011008&NEREL&1410&217\\
&TIME&C0040223&NEREL&128&40\\
A2.1.3 Quantitative&NUMBER&C0237753&NEREL&5202&724\\
&PERCENT&C0439165&NEREL&1948&175\\
&AGE&C0001779&NEREL&505&59\\
A2.1.4 Functional&ADMINISTRATION\_ROUTE&C0013153&NEREL-BIO&295&58\\
A2.1.5.4 Geographical areas&LOCATION&C0450429&NEREL&166&0\\
&COUNTRY&C0454664&NEREL&473&20\\
&STATE\_OR\_PROVINCE&C1555317&NEREL&81&0\\
&DISTRICT&C5447118&NEREL&17&0\\
&CITY&C0008848&NEREL&104&11\\
&FACILITY&C1547538&NEREL&214&36\\
A2.2 Finding&FINDING&&NEREL-BIO&2965&709\\
A2.4 Intellectual product&LAW&C1947938&NEREL&10&0\\
&ORDINAL&C0439080&NEREL&1203&61\\
&PENALTY&C0680633&NEREL&0&0\\
A2.7 Organizations&ORGANIZATION&C1561598&NEREL&531&18\\
A2.8 Group Attribute &NATIONALITY&C0027473&NEREL&25&1\\
A2.9 Group&&&\\
A2.9.1 Professional Group&PROFESSION&C0028811&NEREL&237&22\\
A2.9.3 Family Group&FAMILY&C0015576&NEREL&30&0\\
\midrule
B \textbf{Event}&EVENT&C0441471&NEREL&43&4\\
B1 \textbf{Activity}&&&&\\
B1.1.2 Individual Behaviour&ACTIVITY&C0004927&NEREL-BIO&424&32\\

B1.3.2 Research Activity &SCIPROC&C0242481&NEREL-BIO& 1835&519\\
B1.3.1.1 Laboratory Procedure&LABPROC&C0022885&NEREL-BIO&1644&308\\
B1.3.1.3 Therapy of Preventive procedure&MEDPROC&C0199171&NEREL-BIO&4204&458\\
&HEALTH\_CARE\_ACTIVITY&C0086388&NEREL-BIO&394&28\\
\midrule
B2 \textbf{Phenomenon or Process}&&&&\\
B2.1 Human-caused Phnomenon&CRIME&C0010325& NEREL&1&0\\
B2.2.1.2.1 Disease or Syndrome & DISO& C0012634&NEREL&11576&2020\\
B2.2.1.1 Physiologic Function&PHYS&C0031842&NEREL-BIO&6769&973\\
B2.2.1.1.1.1 Mental Process&MENTALPROC&C0025361&NEREL-BIO&443&65\\
B2.3 Injury \& Poisoning&INJURY\_POISONING&C3263723&NEREL&683&115\\ \bottomrule
\end{tabular}}{}
\end{table*}

NEREL-BIO extends the annotation scheme of the general-domain Russian dataset NEREL \citep{NEREL}. Since NEREL-BIO includes all entity types and relation types from NEREL, we provide a short description of 29 entity types from the NEREL in Sec. \ref{sec:nerel}. In Sec. \ref{sec:nerel-bio}, we describe 17  biomedical entity types that have been utilized in NEREL-BIO (including DISO entity, which is renamed DISEASE  from the general NEREL dataset).

%Note that this paper covers only named entities. Biomedical texts contain numerous nested mentions of entities such as anatomical parts within each other, diseases containing body parts or chemicals, procedures containing names of diseases or devices, etc.

\subsection{Overview of NEREL nested named entities}\label{sec:nerel}
To the best of our knowledge,
% We provide a brief introduction of NEREL here. 
NEREL is the first dataset annotated simultaneously with nested entities, relations between those entities, and knowledge base links \citep{NEREL,NEREL2022}. 
Entity linking annotations leverage nested named entities, and each nested named entity can be linked to a separate Wikidata entity. Nested entities enhance coverage of annotated entities, as well as relations between entities and knowledge base links. For example, using single ORGANIZATION  entity in \textit{Lomonosov Moscow State University} leads to the loss of two internal named entities: CITY (\textit{Moscow}) and PERSON (\textit{Lomonosov}).  In such cases, NEREL employs the nested annotation \textit{Lomonosov\textsubscript{PERSON} Moscow\textsubscript{CITY} State University}.
%As we could see in experiments with NEREL, annotation of nested entities in contrast to so-called ``flat'' entities enhances coverage of annotated entities, as well as relations between entities and knowledge base links. For example,   only one external (flat) ORGANIZATION  entity should be extracted in \textit{Lomonosov Moscow State University}, which leads to the loss of two internal named entities: CITY (\textit{Moscow}) and PERSON (\textit{Lomonosov}). 
% Annotating  internal entities, we also obtain an opportunity to label `local' relations. 
Annotation of internal entities allows `local' relations between nested entities. In the above example, it allows for establishing a relation between the university and its headquarters (\textit{Moscow}). Annotated nested entities enables wider coverage of  entity linking into a knowledge base. For example, entity \textit {Mayor of Novosibirsk} is absent in Wikidata (which is a reference knowledge base for NEREL), but nesting permits linking of the internal entities \textit {Mayor} and \textit {Novosibirsk}. 
%Having nested entities, we can provide a more precise set of connections between texts and a knowledge base.

%We decided to leverage our experience in nested entity annotation in the general-domain NEREL dataset and to apply nested entity annotation to the biomedical domain.  In biomedical annotation, we allowed to use all available entity types from NEREL suggesting that this decision can provide interesting highlights such as differences in nestedness structure in general and biomedical domains and a basis for domain transfer experiments. REPEATS.

%At the time of writing, NEREL is the largest Russian dataset annotated with entities and relations. 
% when compared to the existing Russian datasets. 
At the time of this writing, NEREL contains 56K named entities and 39K relations annotated in 900+ person-oriented news articles. The entity types in NEREL can be categorized as follows: 
\begin{itemize}
\item basic entity types: PERSON, ORGANIZATION, LOCATION, FACILITY and geopolitical entities subdivided into COUNTRY,   CITY, DISTRICT and  STATE\_OR\_PROVINCE  entities;
\item numerical entities: NUMBER, ORDINAL, DATE, TIME, PERCENT, MONEY, AGE; 
\item socio-political entities (NATIONALTY, RELIGION, IDEOLOGY) and LANGUAGE; 
\item law-related entities (LAW, CRIME, PENALTY);
\item work-related entities  (PROFESSION, WORK\_OF\_ART, PRODUCT, AWARD);
\item  DISEASE;
\item  EVENT.
\end{itemize}

The DISEASE entity type is most relevant to the biomedical domain. It can be also noted that PERSON, ORGANIZATION, LOCATION entities from the basic entity group in the general domain were annotated in the biomedical MedMentions corpus \citep{mohan2018medmentions}, locations were also annotated in the QUAERO corpus \citep{neveol2014quaero}. The NEREL profession type corresponds to the MedMentions occupations (OCCU) type. Some NEREL entities (such as WORK\_OF\_ART, AWARD, PENALTY) are less
relevant to the biomedical domain.

\subsection{NEREL-BIO}

\subsubsection{Text Collection}
 
We used sourced documents from the WMT-2020 Biomedical Translation Task collection~\citep{bawden-etal-2020-findings} that contains 6,029 Medline abstracts in Russian and their English translations.\footnote{\url{https://github.com/biomedical-translation-corpora/corpora}} We selected texts in the range of 6--20 sentences. Second, we trained a mBERT-based~\citep{devlin2018bert} NER model on MedMentions~\citep{mohan2018medmentions} and applied it to Russain abstract in a zero-shot fashion. We picked about 100 documents with the densest and most diverse recognized entities. Based on the analysis of this automatic annotation, we %decided that  formulated basic principles for abstract inclusion into the NEREL-BIO corpus: 
selected abstracts with disease mentions and related laboratory or medical procedures for including in NEREL-BIO.% --- \textcolor{red}{UNCLEAR rephrase last sentence} 

% The subsequent selection consisted of two stages. At the first stage about 100 documents were extracted with maximal density of diverse entity types in a document. With this aim, the Multilingual BERT model \citep{devlin2018bert} trained on the MedMentions corpus \citep{mohan2018medmentions} was applied to Russian MEDLINE abstracts. These documents allowed us to formulate basic principles of abstract selection for the NEREL-BIO corpus.\pb{??} At this stage, parallel English abstracts were also extracted and later annotated. 
% At the second stage, abstracts with disease mentions and related laboratory or medical procedures were selected to NEREL-BIO. 

%The abstracts were annotated using the BRAT annotation tool \citep{stenetorp2012brat}. To facilitate manual annotation, the dataset was automatically labeled with two models: multilingual BERT \citep{devlin2018bert} trained on the English MedMentions \citep{mohan2018medmentions} for biomedical entity recognition (10 entity types) and MRC model \citep{li2020unified} trained on the NEREL dataset, which allowed labeling nested entities from the general domain (29 entity types).

The abstracts were annotated using the BRAT annotation tool \citep{stenetorp2012brat}. To facilitate manual annotation, initial annotation was done automatically with two models: multilingual BERT \citep{devlin2018bert} trained on the English MedMentions \citep{mohan2018medmentions} for biomedical entity recognition (10 entity types) and MRC model \citep{li2020unified} trained on the NEREL dataset, that helped labeling nested entities from the general domain (29 entity types). The automatic techniques provided the annotation of most evident entities and became a basis for further manual labeling.
%\textcolor{red}{Do you mean initial annotation was done automatically and then corrected manually. I assumed this to be true. The purpose of automatic annotation needs to be made clear}

Table~\ref{tab:generalstats} summarizes statistics of NEREL-BIO in terms of documents and entity mentions.
% presents the sizes of the Russian and English parts of NEREL-BIO in documents and mentions. 
Table~\ref{tab:diseases} contains most frequent disease mentions in the Russian part of NEREL-BIO. It can be seen that abstracts are quite diverse in content.

\begin{table*}[t]
\centering
\processtable{UMLS semantic groups and NEREL-BIO entity types\label{tab:entity_examples}}
{\begin{tabular}{@{}l|l|l@{}}\toprule

%\hline
\textbf{NEREL-BIO entity} & \textbf{Explanation} 
&\textbf{Examples in the dataset (translated from Russian)}\\
\midrule
ACTIVITY&human or animal behavior&smoking, contact with animals, to give up smoking,\\
&&military actions, illegal drug use\\
\midrule
ADMINISTRATION\_ROUTE&ways of administering a drug  or another chemical to an organism for &enteral, rectal, intranasal, intravenous, inhalation\\
&treatment including drug forms&injection, infusion\\
\midrule
ANATOMY&comprises organs, body part, cells and cell components, &eye, bone, brain, lower limb, oral cavity, blood, \\
&body substances&anterior lens capsule, right ventricle, lymphocyte\\
\midrule
CHEM&chemicals including legal and illegal drugs, biological &opioid, lipoprotein, iodine, adrenalin, memantine,\\
&molecules&methylprednisolone\\
\midrule
DEVICE&manufactured objects used for medical purposes&catheter, prosthesis, tonometer, tomograph\\
&&removable prosthesis, stent, metal stent\\
\midrule
DISO&any deviations from normal state of organism: diseases, symptoms,  &appendicitis, haemorrhoids, magnesium deficiency\\
&dysfunctions, abnormality of organ, excluding injuries or&deep vein thrombosis, Diabetes Mellitus,pain\\
&   poisoning&spine pain, complication, bone cyst, acute inflammation\\
\midrule
FINDING&does not have direct correspondence in UMLS, it conveys&longer hospital stay, stopped the progression of \\
&the results of scientific study described in the abstract&keratoconus, stabilize the glaucoma process\\
\midrule
FOOD&Substances taken in by the body to provide nourishment&salt, milk, hot meal\\
&including fresh water, alcoholic beverages&breast  milk (in context of breastfeeding)\\
\midrule
GENE&nucleic acid sequences that function as units of heredity&KIR gene, PARK2, RASSF1A, allele, CYP2C19\\
\midrule
HEALTH\_CARE\_ACTIVITY&health care administration and organization activities &hospitalization, discharge from hospital, medical evacuation\\
\midrule
INJURY\_POISONING&damage inflicted on the body as the direct or indirect result &overdosing, burn, drowning,\\
& of external force including poisoning&falling, childhood trauma\\
\midrule
LABPROC&testing body substances and other  diagnostic procedures &biochemical analysis, polymerase chain reaction test,\\
&such as ultrasonography&electrocardiogram, histological\\
\midrule
LIVB&Any individual living (or previously living) being&rat,mosquito, mouse, dog, parasite,\\
&except humans&rabbit, virus\\
\midrule
MEDPROC&procedures concerned with remedial treatment of diseases,&lung resection, antiviral therapy, eyelid surgery, \\
&including surgical procedures&chemotherapy, corneal transplant,appendectomy\\
\midrule
MENTALPROC&conceptual functions or thunking&self-esteem,psycho-emotional status, cognitive function\\
\midrule
PHYS&biological function or process in organism including organism&blood flow, childbirth, uterine contraction,\\
& attribute (temperature) and excluding mental processes&arterial pressure, body temperature\\
\midrule
SCIPROC&scientific studies including mathematical methods or &retrospective analysis, confidence interval, ICD-10,\\
& clinical studies, scales, classifiers, etc.&analysis of variance, multivariate analysis\\

\botrule
\end{tabular}}{}
\end{table*}

\subsubsection{Entity Types}\label{sec:nerel-bio}
%As other researchers working with biomedical abstracts, 
Biomedical entity types for selected for annotation based on their presence in the UMLS taxonomy and other annotated datasets in the biomedical domain. 16 specialized biomedical entity types and 29 entity types from the general NEREL dataset are included in NEREL-Bio. The full set of entity types, explanations, and examples in NEREL-BIO are presented in Table~\ref{tab:entity_examples}.

%To choose biomedical entity types for annotation, we started with the analysis of the UMLS taxonomy and other annotated datasets in the biomedical domain. As a result, we added 16 specialized biomedical entity types to 29 entity types of the general NEREL dataset. The full set of entity types, explanations, and examples in NEREL-BIO are presented in Table~\ref{tab:entity_examples}.

Biomedical entity types  in Table~\ref{tab:entity_examples} correspond to the most relevant UMLS concepts and are annotated according to 
%given definitions based on the
UMLS definitions. There are a few exceptions as given below:
\begin{itemize}
    \item HEALTH\_CARE\_ACTIVITY, which is described as a quite general concept in UMLS, is treated as health care administration and organization activities such as \textit{hospitalization} or \textit{medical evacuation}; 
    \item LABPROC entity comprises both laboratory and other diagnostic procedures;
    \item FINDING entity does not have a direct correspondence in UMLS, it conveys 
the results of the scientific study described in the abstract, e.g. \textit{longer hospital stay, stopped the progression}.
\end{itemize}

NEREL-BIO entity types in Table~\ref{tab:entity_types} are attached to the most relevant UMLS semantic types and ordered according to the UMLS taxonomy. Also for each entity type, the corresponding UMLS concept is found and its identifier (CUI) is included in the table. Table~\ref{tab:entity_types} also contains statistics for the Russian and English parts of the NEREL-BIO corpus. 

It can be seen that all entity types were successfully linked to the UMLS taxonomy except for WORK\_OF\_ART which is missing in UMLS.
%which does not have a direct corresponding concept in UMLS and has been never met in the dataset. 
Rare or absent entity types in the NEREL-BIO dataset are as follows: IDEOLOGY (0), RELIGION (0), AWARD (2), LANGUAGE (5), PENALTY (0), CRIME (1), and LAW (10). At the same time, we could see quite diverse mentions of geographical locations and some of the money (mainly in the context of medical expenses). Mentions of professions or occupations are quite frequent: mainly medical specialists are mentioned, but also there are studies on occupational diseases of specific professional groups.

Some principles of annotation employed in the general domain were changed NEREL-BIO. In particular, in the general domain, mainly capitalized mentions were annotated as named entities. In the biomedical domain, the same entity types can also appear as lower-cased mentions:
\begin{itemize}
\item any humans or groups can be annotated with the label PERSON  such as \textit{patient}, \textit{control group}, \textit{population with low income};
\item ORGANIZATION tag is used not only for tagging specific organizations but organization types such as \textit{hospital, medical institution, rehabilitation center}.
\item location-related tags (LOCATION, COUNTRY, CITY, STATE\_OR\_PRO-VINCE, DISTRICT, FACILITY) are also used in both cases:\textit{rural settlement, low-income countries, coastal areas}.
\end{itemize}

%Annotated entities can be single words (nouns, verbs, or adjectives) or multi-word expressions (usually noun groups). Entities can be nested and embedded in each other. For example, the medical procedure   \textit{maintenance therapy with opioid agonists} (PTAO) (translated from Russian) contains a mention of a chemical. REPEATS

%Annotation of nested entities in the biomedical domain is more difficult than in the general domain because in the general domain most entities are usually capitalized, so in labeling, we can rely on the natural capitalization of names. %\pb{repetition? see above} 

%In the biomedical domain, most entities are written lowercase. At the same time annotation of nested entities has some advantages. In particular, we can annotate some specific entity (a specific variant of a disease), which is absent in any referential resource such as UMLS.  This further will allow labeling relations for this specific entity. Additionally,  we can annotate an internal, more general entity, which can be directly linked to some known concept. %\pb{rewrite one of the \textbf{at the same time}}

Entities in NEREL-BIO often appear lower-cased while being absent in UMLS. For example, the term \textit{left-sided congenital diaphragmatic hernia} is absent in UMLS. We annotate this as follows:

$[left-sided \quad  [congenital \quad  [[diaphragmatic]_{ANATOMY}$  

$[hernia]_{DISO}]_{DISO}]_{DISO}]_{DISO}$

 Although  we cannot link the whole term in UMLS, we can link the sub-terms: Hernia (C0019270), Diaphragmatic Hernia (C0019284), Respiratory Diaphragm (C0011980), Congenital diaphragmatic hernia (C0235833).  
 %Internal annotated entities, which have correspondences in UMLS,  can be linked to their relevant concepts.%\pb{example?}

For annotating multiword terms we followed the following guidelines: % are as follows:
\begin{itemize}
\item  two-three word terms in form of noun groups without prepositions discussed in texts  are annotated without additional checks;
\item  longer multiword phrases containing prepositions should be supported with some additional evidence, for example, there can be an abbreviation in the text for a long multiword term (\textit{ST-segment elevation acute coronary syndrome -- STSEACS}), a long term or its English equivalent can be found in UMLS (\textit {Metastasis from malignant tumor of liver} C1282502) or other biomedical resources;
\item internal spans in an annotated multiword term (single words or phrases), which can be considered to be valid biomedical terms, are also annotated with corresponding entity types;
\item general adjectives, adjectival quantifiers are not included in the annotated entity: \textit{various tumors} are annotated as 
$various [tumors]_{DISO}$.

\end{itemize}
%% statistics of nested entities

The annotation scheme was created during multi-round preliminary annotation of parallel Russian and English abstracts. Terminologists experienced in terminological studies including the  biomedical domain were involved in the annotation. All annotated abstracts were additionally checked by a moderator.

% Text:
In Table \ref{tab:nested_stat} we provide a brief summary of how frequently nested entities appear in NEREL-BIO. For each entity type, we counted how many times entities of this type appear as an outer entity (eliminating multiple occurrences of the same entity), and divide this number by the total occurrences of the entity type in the corpus. Then we filter out the types with less than 200 occurrences in the corpus. The top ten entity types along with their nestedness frequency are presented in the table. Frequencies in the parallel English / Russian abstracts of the NEREL-BIO are shown in the last two columns of Table \ref{tab:nested_stat}. Here we compare only 100 parallel abstracts for each language.

Frequencies of nested entities in Russian and the smaller English corpus were mostly comparable. The differences can be explained by the following:  1. the abstracts are not fully parallel: paper titles are absent in Russian abstracts but included in English abstracts; 2. the different syntax of languages determines different structures of sentences and nestedness; 3. sentences in Russian and in English are not always direct translations but can be significantly reformulated. Two last factors especially affected the FINDING entity since these can be long and therefore can be formulated in multiple ways.

Additionally, we analyzed nested entities in the following way. We aggregate typical pairs of nested entities from the corpus. Each pair has an outer and an internal entity. Table \ref{tab:nested_stat2} presents top ten pairs of types for such entities. Note, that an outer entity can contain one, two or more internal entities. 
In fact, the NEREL-BIO dataset has outer entities that contain up to eight internal entities at the same level of nestedness.

% For example, FINDING entity with text span \begin{otherlanguage}{russian}`достоверно снижались количество эпителиоцитов десны TLR-2 и TL-4, популяции CD4- и CD8-лимфоцитов воспалительного инфильтрата'.\end{otherlanguage}. 
Therefore, we provide raw counts in the table. Overall, the Russian part of the NEREL-BIO contains 22,392 such pairs (the English subset has 3,864 nested entity pairs).

% tables:

\begin{table}
\centering
\processtable{Frequencies of top ten entity types with nested entities in full Russian collection and 100 Russian and English documents for comparison. \label{tab:nested_stat}}
{\begin{tabular}{l r r r}\toprule
	\textbf{Entity Type} & 	 \textbf{Full RU, in \%} & 	 \textbf{EN, in \%} & 	 \textbf{RU, in \%}  \\
	\midrule

FINDING	& 	65.7        & 71.2 & 57.4 \\
PHYS	& 	38.3        & 40.7 & 39.8 \\
INJURY\_POISONING&37.7  & 49.0 & 39.4 \\
DISO & 		37.3        & 41.2 & 37.6 \\ 
DEVICE	& 	33.9        & 42.5 & 46.2 \\  
LABPROC	& 	30.2        & 34.8 & 31.0 \\
MEDPROC	& 	30.0        & 44.7 & 33.1 \\
ANATOMY	& 	27.3        & 28.3 & 31.0 \\
SCIPROC	& 	23.9        & 32.1 & 24.6 \\
CHEM	& 	22.5        & 20.1 & 17.2 \\
\botrule
Total Entities & 66,286	 & 9,961 & 9,209 \\
\botrule
Total (Outer) Nested  & 17,182 & 3,002 & 2,468 \\
Entities&&&\\
\end{tabular}}{}
\end{table}

% total
% ent_type	freq
% 15	FINDING	0.656526
% 32	PHYS	0.383455
% 19	INJURY_POISONING	0.376855
% 10	DISO	0.372806
% 9	DEVICE	0.339422
% 20	LABPROC	0.301686
% 24	MEDPROC	0.299523
% 3	ANATOMY	0.273277
% 35	SCIPROC	0.238789
% 5	CHEM	0.224543

% English

% ent_type	freq
% 10	FINDING	0.712146
% 9	FACILITY	0.527778
% 12	INJURY_POISONING	0.490000
% 15	MEDPROC	0.446903
% 6	DEVICE	0.425000
% 7	DISO	0.411918
% 20	PHYS	0.407281
% 13	LABPROC	0.347518
% 23	SCIPROC	0.320961
% 0	ACTIVITY	0.312500
% 2	ANATOMY	0.283343
% 14	LIVB	0.209677
% 3	CHEM	0.201183
% 5	DATE	0.133641
% 19	PERSON	0.130636

% Russian

% 	ent_type	freq
% 9	FINDING	0.574176
% 5	DEVICE	0.461538
% 19	PHYS	0.398281
% 11	INJURY_POISONING	0.393939
% 17	ORGANIZATION	0.382353
% 6	DISO	0.375991
% 8	FACILITY	0.368421
% 14	MEDPROC	0.331169
% 0	ACTIVITY	0.315789
% 12	LABPROC	0.310219
% 2	ANATOMY	0.310082
% 22	SCIPROC	0.245763
% 15	MENTALPROC	0.214286
% 3	CHEM	0.171792
% 18	PERSON	0.162791

\begin{table}
\centering
\processtable{Top ten nested entity pairs in NEREL-BIO.\label{tab:nested_stat2}}
{\begin{tabular}{l l r}\toprule
	\textbf{Outer Entity Type} & 	\textbf{Internal Entity Type} & 	 \textbf{Occurrences} \\
	\midrule

DISO & 	DISO & 	3,380  \\
ANATOMY & 	ANATOMY & 	3,051  \\
DISO & 	ANATOMY & 	1,476  \\
PHYS & 	PHYS & 	1,267  \\
CHEM & 	CHEM & 	1,116  \\
PERSON &	PERSON &	1,038 \\
FINDING & 	PHYS & 	956  \\
MEDPROC &	MEDPROC &	911 \\
PHYS & 	ANATOMY & 	786  \\
PHYS & 	CHEM & 	523  \\
\botrule
Total Nested Pairs & & 22,392\\
\botrule
\end{tabular}}{}
\end{table}

% 79	DISO	DISO	3380
% 23	ANATOMY	ANATOMY	3051
% 76	DISO	ANATOMY	1476
% 311	PHYS	PHYS	1267
% 38	CHEM	CHEM	1116
% 286	PERSON	PERSON	1038
% 141	FINDING	PHYS	956
% 228	MEDPROC	MEDPROC	911
% 292	PHYS	ANATOMY	786
% 293	PHYS	CHEM	523

% outer_type	inner_type	filename
% 80	DISO	DISO	3976
% 24	ANATOMY	ANATOMY	3599
% 77	DISO	ANATOMY	2464
% 320	PHYS	PHYS	1449
% 39	CHEM	CHEM	1286
% 142	FINDING	PHYS	1270
% 300	PHYS	ANATOMY	1165
% 294	PERSON	PERSON	1058
% 232	MEDPROC	MEDPROC	1034
% 301	PHYS	CHEM	743

%% statistics of nested entities

\section{Experiments and Evaluation}
For our experiments, we split NEREL-BIO into train/dev/test subsets (612/77/77 documents). For name-entity recognition experiments, we report results (see Table~\ref{tab:generalstats}) on (i) a Machine Reading Comprehension (MRC) model~\citep{li2020unified}\footnote{Our code is available at \url{https://github.com/fulstock/mrc_nested_ner_ru}}.; and (ii) a sequence model~\citep{shibuya2020nested}. 
%We perform experiments on the  Russian part of the NEREL-BIO dataset evaluating results on 32 entity types having frequencies  more than 50 mentions in the dataset.

\subsection{Models}
MRC task is formulated in the following way: for the given context $X$ and question $Q$ the model should obtain answer $A$ with some function $F$ defined as $A = F(X, Q)$. In the named entity recognition task, $X$ would be the given sentence/paragraph; $Q$ is some generated or selected query sentence for a given named entity type; $A$ is the subsequence of the context $X$ that denotes the named entity; $F$ is the retrieving model itself.

For the sequence model, we employed three binary classifiers based on the output of the last hidden layer from the RuBERT model (Russian BERT)  \citep{kuratov2019rubert}. The first classifier determines the starting position of the named entity. The second classifier determines the ending position of a named entity (possibly different) of the same class. The third classifier decides, whether chosen start-end pairs represent a single named entity of such class. These classifiers are trained for each class (type) separately. 
% In such way it is possible to retrieve nested named entities. 
%For the MRC model, we employ RuBERT \citep{kuratov2019rubert} as a basis for the MRC model. 
Batch size was set to 16 with maximum length of the sequence to be 192 tokens. Model was trained during 16 epochs on 8 Tesla V100 GPUs. Other parameters set to default values after ~\citep{li2020unified}.

For training the MRC model, we each entity type (e.g. ORGANIZATION), we employed manually collected definitions of corresponding concepts from dictionaries (including Wikipedia), frequent mentions of an entity type in the training collection, some contexts (sentences) from the training collection, and keywords (e.g. ORG).

%Questions for NER may be keywords (\texttt{ORG} $\mapsto$ \textit{organization}), definitions of corresponding concepts in some dictionaries (for example, in Wikipedia), most frequent examples of an entity in the training collection, some contexts (sentences) from the training collection, etc.~\citep{li2020unified,zhou2021improved,dongho2021demonstration,rozhkov2022mrc}.  

We compared several question variants:

\textbf{Keyword}: the question consists of entity tags such as DISO or ANATOMY ~\citep{li2020unified}\

 \textbf {Component-based}: 2-5-10 most frequent lemmatized components of a given entity are used for formulating a query, for example ``DISO are entities such as a tumor, complication, disorder, disease, illness'' (5-component example). Previous experiments with the general NEREL dataset showed that component-based questions outperformed other variants~\citep{rozhkov2022mrc};
 
\textbf{Contextual}: a sentence from the training sample containing a named entity of a given type without explicit or implicit labeling used for this entity in the sentence. For example, a question for DISO entity type can be as follows: ``60 patients in the most acute period of hemispheric ischemic stroke were examined.''

\textbf{Lexical}: as in the contextual variant, a sentence from the training corpus is used as a question; additionally, the entity of a given type is masked with its label \citep{zhou2021improved}. An example of a masking sentence with several mentions of an entity looks as follows. The initial sentence contains three mentions of DISO: ``The addition of gout contributes to endothelial dysfunction and worsens the course of hypertension.''. The corresponding 
lexical question: ``The addition of DISO contributes to DISO and worsens the course of DISO.''

We used the so-called full lexical approach, when all entities in a sentence of a given type are substituted with masks. If a longer entity contains a shorter entity of the same type, the longer entity is preferred (outmost variant). The example of the lexical variant corresponding to the above-mentioned contextual example is as follows: ``60 patients in the most acute period of hemispheric DISO were examined''.

The selection of a  sentence for contextual or lexical questions is carried out in the following manner:
\begin{itemize}
    \item The most frequent entity for a given entity type is selected
    \item The first sentence in the training set that contains the selected entity is extracted to be used as a question. By ``first'' we imply here the lexicographic order of the filenames of the original dataset. 
\end{itemize}

We also provide experimental results for the \textit{second-best Sequence model}~\citep{shibuya2020nested} since it gave comparable results in the NEREL dataset. For this setup, we employed RuBERT model with batch size set to 16 and the same length of 192 tokens. The model was trained for 32 epochs on 8 GPUs while other parameters were set to default values.  

%The model treats the tag sequence for nested entities as the second best path within the span of their parent entity. In addition, the decoding method for inference extracts entities iteratively from outermost ones to internal ones in an outside-to-inside way. It uses the Conditional Random Field method as an output layer. 

\subsection{Results}
Span-level micro- and macro-averaged precision, recall, and F1 results of the models are shown in Table \ref{tab:results}. The performance of the 5-component MRC model for the ten most frequent entities is presented in Table~\ref{tab:results_entity}. 

As shown in Table \ref{tab:results}, the best macro-averaged results are achieved by the 5-component model. Depending on entity types, performance of the 5-component model varies greatly (see Table~\ref{tab:results_entity}). In particular, this model achieves 85\% F1 and 61\% on ANATOMY and PHYS, respectively. We note that the best obtained results of nested NER for NEREL-BIO are lower than for NEREL (where MRC model achieved 80\% micro-F-measure). This is in line with existing published NER results obtained that also show similar decreased results on biomedical texts \citep{shibuya2020nested,liu2022handling}. The results for second-best sequence model are closest  to the MRC model in micro measures but  significantly worse in macro measures. This can be partly be explained by the low amount of training data for specific entity types \citep{Artemova2022runne}.

\begin{table}[t]
\centering
\processtable{Results of nested NER models on NEREL-BIO. \label{tab:results}} 
{\begin{tabular}{@{}lllll@{}}
\hline
\textbf{Model} & \textbf{Precision} & \textbf{Recall}& \textbf{MICRO-F}&\textbf{MACRO-F}\\
\hline
MRC&&&&\\
Keyword&\textbf{77.72}&75.92&76.81&61.80\\
2-comp&66.83&59.90&\textbf{76.94}&61.94\\
5-comp&77.64&75.85&76.74&\textbf{62.25}\\
10-comp&77.23&\textbf{76.52}&76.87&61.65\\
Lexical&77.33&75.41&76.36&61.37\\
Contextual&73.09&76.08&74.54&59.63\\
\midrule
Second-best&75.28&72.98&74.10&51.29\\
\hline
\end{tabular}}{}
\end{table}

\begin{table}[t]
\centering
\processtable{Results of the 5-component MRC model on most frequent entity types. \label{tab:results_entity}} 
{\begin{tabular}{@{}llll@{}}
\hline
\textbf{Model} & \textbf{Precision} & \textbf{Recall}& \textbf{F1}\\
\hline
ANATOMY&83.31&86.75&85.00\\
CHEM&80.56&83.38&81.94\\
DATE&76.88&78.25&77.56\\
DISO&78.96&82.03&80.47\\
LABPROC&69.77&62.50&65.94\\
MEDPROC&70.61&78.44&74.32\\
NUMBER&86.36&90.20&88.24\\
PERCENT&94.50&93.99&94.24\\
PERSON&84.17&93.85&88.74\\
PHYS&57.72&64.59&60.96\\
\hline
\end{tabular}}{}
\end{table}

\subsection{Error Analysis}
%\textcolor{red}{Say how the errors were spotted and chosen for error analysis. Also, there is no error analysis for sequence model.}
We analysed the results of the best MRC model on the NEREL-BIO test collection in comparison with manual annotation and  found the following frequent types of errors:
\begin{itemize}
    \item misclassification of abbreviations, which can be of different entity types but look very similar: IPN (iskrivliniye peregorodki nosa -- deviated septum of the nose), MPT (methadone maintenance therapy). 
    \item {evidently longer entities than necessary were extracted: including verbs (``subgroup was taken in the 2nd group''), including conjunctions (``ART and MMT''), etc.}, with a comma in the middle (``EMBASE, Medline''), etc.;
    \item some irrelevant entities can be labeled, for example, ``level of education'' was classified as  PHYS.
\end{itemize}
Missing entities were found in human markup due to the difficulty of the annotation task itself. 
For example, in the ``mild cognitive impairment'' phrase, an annotator missed labeling ``cognitive impairment''.

\section{Discussion and Limitations}
Several issues may potentially limit the applicability of NEREL-BIO; they are mostly shared with other available datasets.

\paragraph{Seen and unseen mentions of entities.} Recent works on BERT-based models for information extraction demonstrate that the generalization ability of these models is influenced by domain shift or whether the test entity/relation has been seen in the training set \cite{miftahutdinov2020biomedical,tutubalina2020fair,kim2022your}.  To avoid such biases, \cite{kim2022your} removes overlaps in entity mentions and concept identifiers between training and test sets while \cite{tutubalina2020fair} focuses on zero-shot entity linking between different concept terminologies. %\textcolor{red}{Unclear how zero shot EL is related to the previous point.} 
We leave these approaches to future work. We plan to investigate how well MRC models for nested NER can be adapted to unseen mentions. \\

%\textcolor{red}{Need to say something about annotation guidelines and how they were used. Also saying something about inter-annotator agreement is a good thing.}\\

%We plan to investigate how well MRC models for nested NER can be adapted to unseen mentions. 
%In this study, we use the random train/test split, leaving more advanced split options for future work.
%
\paragraph{Knowledge transfer between general and biomedical domains.} The proposed NEREL-BIO corpus shared annotation scheme with our general-domain dataset NEREL for common entity types such as AGE, NUMBER, FACILITY, and ORGANIZATION (29 types in total). Transferability of trained models across two datasets with completely different contexts can be limited due to domain shift, while sequential training can cause complete retraining of model weights. We mark the investigation of strategies for combining different domains for future work.

\paragraph{Disease-centric abstracts} NEREL-BIO includes PubMed abstracts describing the results of clinical trials, hospitalization, and treatment of patients. The most frequent entities (e.g., diseases, injury, and anatomy) are related to a clinical domain, while biological entities such as genes and proteins are less presented. We suppose that this restricts the extraction of new biological relationships for protein-protein interaction or knowledge graph completion tasks, which will require additional data annotation.

\section{Conclusion}
 Biomedical texts contain numerous nested mentions of entities such as anatomical parts within each other, diseases containing body parts or chemicals, names of procedures, which include diseases or devices, etc. In this paper, we presented the first Russian dataset of biomedical abstracts NEREL-BIO, annotated with nested entities. The selected abstracts focus primarily on diseases and related medical procedures. The dataset contains a small collection of annotated parallel English abstracts. Our annotation shows that nested entities provides a better basis for extracting relations that would otherwise be lost. Similarly, nested entities also permit more complete entity linking to knowledge bases. Since, NEREL-BIO extends the annotation scheme of the general-domain Russian NEREL dataset, it permits studying domain transfer methods.

%Current work-in-progress is the annotation of relations between entities annotated in the NEREL-BIO dataset as well as linking entities to UMLS concepts, which allows the creation of the Russian biomedical dataset of disease-related scientific abstracts with three levels of annotations over nested entities.

\section*{Acknowledgements}
The project is supported by the Russian Science Foundation, grant N 20-11-20166.
The authors thank all annotators for their contribution and Zulfat Miftahutdinov for his work on models trained on the MedMentions corpus.

\bibliographystyle{natbib}
\bibliography{ml}

\end{document}